\pdfoutput=1
\PassOptionsToPackage{x11names}{xcolor}
\documentclass[11pt]{article}

\usepackage{ACL2023} 

\usepackage{times}
\usepackage{latexsym}
\newcommand{\cross}[1][1pt]{\ooalign{%
  \rule[1ex]{1ex}{#1}\cr
  \hss\rule{#1}{.7em}\hss\cr}}
\usepackage[T1]{fontenc}

\usepackage[utf8]{inputenc}

\usepackage{inconsolata}
\usepackage{booktabs}
\usepackage{hyperref}
\usepackage{graphicx}
\graphicspath{ {./images/} }
\usepackage{multirow}
\usepackage{ragged2e}
\usepackage{color}
\usepackage{array}
\usepackage{arydshln}
\usepackage{subcaption}
\usepackage{linguex}
\usepackage[switch]{lineno}
\usepackage{tcolorbox}
\usepackage{tablefootnote}

\newcommand{\corpname}{GUMsley}

\title{\corpname{}: Evaluating Entity Salience in Summarization for 12 English Genres}
\author{Jessica Lin \and Amir Zeldes\\
         Department of Linguistics \\ Georgetown University \\ \texttt{\{yl1290, amir.zeldes\}@georgetown.edu}}

\begin{document}
\maketitle
\begin{abstract}
As NLP models become increasingly capable of understanding documents in terms of coherent entities rather than strings, obtaining the most salient entities for each document is not only an important end task in itself but also vital for Information Retrieval (IR) and other downstream applications such as controllable summarization. In this paper, we present and evaluate \corpname{}, the first entity salience dataset covering all named and non-named salient entities for 12 genres of English text, aligned with entity types, Wikification links and full coreference resolution annotations. We promote a strict definition of salience using human summaries and demonstrate high inter-annotator agreement for salience based on whether a source entity is mentioned in the summary. Our evaluation shows poor performance by pre-trained SOTA summarization models and zero-shot LLM prompting in capturing salient entities in generated summaries. We also show that predicting or providing salient entities to several model architectures enhances performance and helps derive higher-quality summaries by alleviating the entity hallucination problem in existing abstractive summarization. 
\end{abstract}

\section{Introduction}
The task of \textit{salient entity extraction} (SEE) is to identify entities that are central to a document's overall meaning. Previous work on SEE has relied on crowdsourcing \cite{dojchinovski2016crowdsourced} or user statistics on the web (e.g.~clickstream data, \citealt{gamon2013identifying}) to derive salience labels for entities. In this study, we extend an approach from \citet{dunietz2014new}, who considered an entity salient if it also appears in a human-written summary or abstract of a news article, and we cover many further genres rather than just news. Figure \ref{fig:sal} shows an example of salient entities in a \textit{conversation} annotated according to our definition of salience. \par
\begin{figure}[h!tb]
\centering
\includegraphics[width=0.5\textwidth]{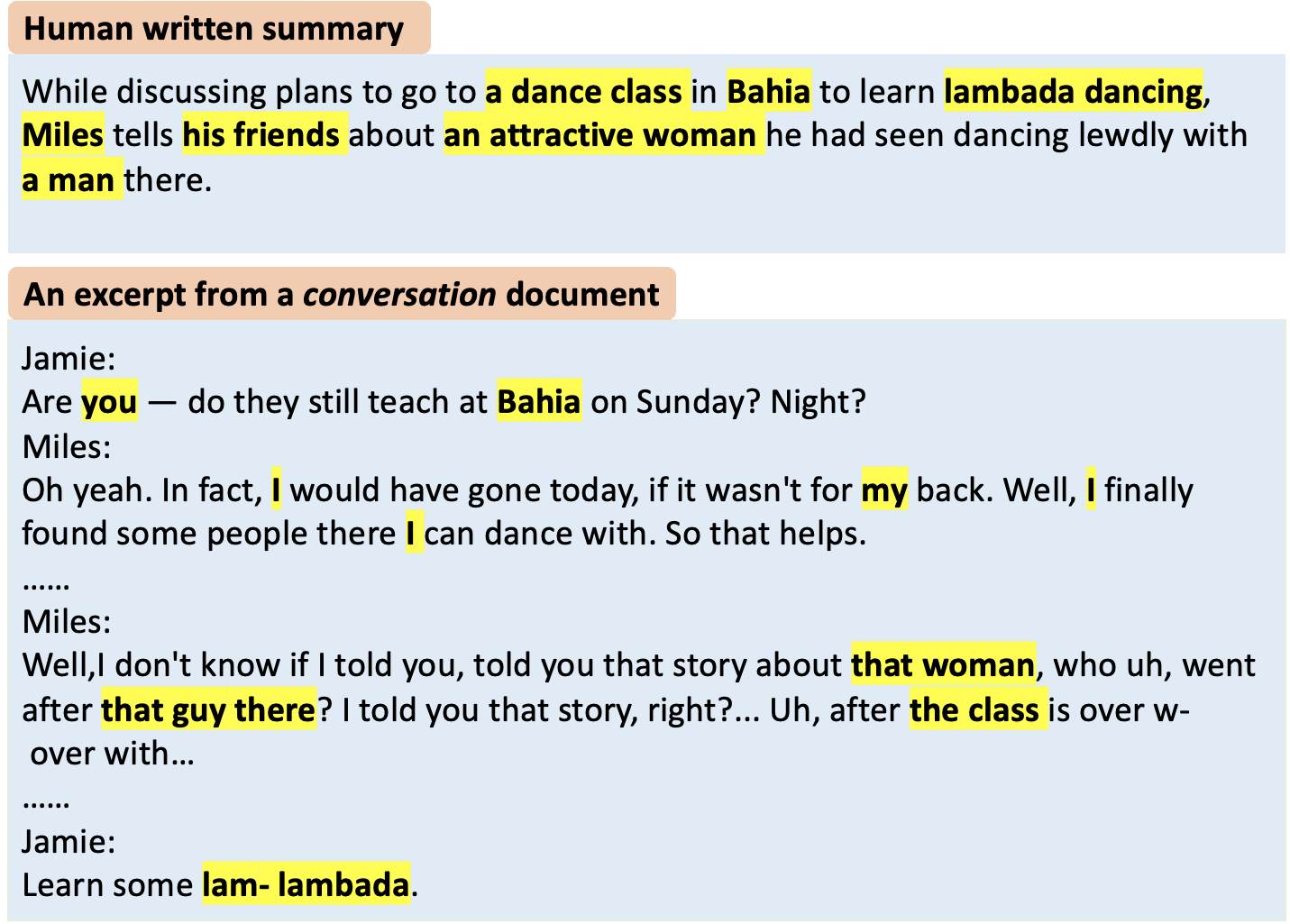}
\par\vspace{-6pt}\par
\caption{A salient entity example from our data. Salient entity mentions are highlighted in yellow.}
\par\vspace{-15pt}\par
\end{figure}\label{fig:sal}

SEE is increasingly important as NLP systems move from understanding document `aboutness' at the word level (e.g.~keyword extraction) \cite{tomokiyo2003language} to entity level document understanding \cite{maddela-etal-2022-entsum, nan-etal-2021-entity}. Therefore, a dataset with SEE labels can benefit downstream applications such as information retrieval and summarization, which extract salient information from large documents and prioritize specific entities in controllable models.

Although several SEE datasets already exist \cite{dojchinovski2016crowdsourced, dunietz2014new, gamon2013identifying, trani2018sel,wu2020wn}, most are predominantly collected from news articles and derive labels using crowdsourcing or ``found'' information such as hyperlinks, which are not intended to annotate salience per se. This has two major limitations: First, crowdsourcing SEE without rigorous training and clear definitions of salience may be biased towards individuals and inconsistent interpretations of what is considered salient. Second, focusing on news limits system performance on more diverse data (e.g.~conversation, vlogs, etc.). 

To investigate the role of SEE in tasks such as summarization, previous entity-centric work \cite{fan-etal-2018-controllable, he-etal-2022-ctrlsum, xiao2022entity} has compared summaries generated by entity-aware methods with generic summarization methods qualitatively. As part of our evaluation, we combine manual and automatic, qualitative and quantitative analyses to assess SEE impact on several approaches to summarization. \par

In this paper, we therefore present and evaluate a gold standard dataset manually annotated with SEE labels, by identifying all entities that appear in a human-written summary as salient, making the task less subjective. Our dataset, called \corpname{} (\textbf{GUM s}alient \textbf{l}inked \textbf{e}ntit\textbf{y} corpus) is based on the existing UD English GUM corpus (Georgetown University Multilayer corpus, \citealt{zeldes2017gum}) and goes beyond other entity salience datasets in covering all named and non-named salient entities for 12 genres of English text. \corpname{} also enables the evaluation of SEE annotations in a broad spectrum of genres and tasks, since the data contains Wikification identifiers for named entities, as well as comprehensive coreference resolution. Our results show that a significant amount of salient entities are not captured by SOTA abstractive summarization models or out-of-the-box LLMs 
(Section \ref{sec:4.1}). We also conduct a quantitative analysis to show that providing gold or even predicted salient entities to models helps to generate a higher quality summary (Section \ref{sec:4.2}). 

\section{Related Work}
\paragraph{Entity Salience Datasets}
The growing interest in SEE is demonstrated by increasing numbers of annotated datasets, with different approaches to recognizing entities and assigning labels. The first step is usually entity identification. While some datasets \cite{dunietz2014new, gamon2013identifying} apply a multi-step NLP pipeline (NP extraction, coreference resolution, possibly a named entity resolver) to pinpoint entities, others \cite{dojchinovski2016crowdsourced,trani2018sel,wu2020wn} have done so manually. Since pipelines may propagate errors to later steps, full manual annotation is used in our study to avoid such issues. To collect salience labels, most studies \cite{dojchinovski2016crowdsourced,trani2018sel} prefer human annotation using crowdsourcing. Although crowdsourcing may outperform automated methods, it is inevitably noisy and can suffer from \textit{subjective bias} issues \cite{maddela-etal-2022-entsum} since people have different judgments on what they consider \textit{salient}. In this study, we follow a more regimented approach similar to the NYT salience corpus (\citealt{dunietz2014new}), which considers entities that also appear in an abstract or summary as salient. Unlike \citeauthor{dunietz2014new}, which uses automatic coreference resolution to detect mentions in newspaper summaries, we use fully manual annotation coupled with GUM's carefully written summaries which have consistent guidelines and style across 12 English genres \cite{LiuZeldes2023}, rather than found abstracts or teasers limited to news or academic data. 

\paragraph{Entity-centric Summarization}
Research on entities in automatic summarization has seen a surge of interest in the NLP community recently. However, numerous studies \cite{cao2018faithful, kryscinski-etal-2019-neural,nan-etal-2021-entity} have pointed out that abstractive summarization models suffer from entity hallucination, i.e.~summaries contain entities that never appear in the source document. Previous attempts to solve such problems include training models to classify whether generated summaries are factually consistent with input documents \cite{kryscinski-etal-2019-neural} and filtering out entities that have no match in the source document \cite{nan-etal-2021-entity,xiao2022entity}. In this study, we propose adapting methods from controllable summarization \cite{fan-etal-2018-controllable, nan-etal-2021-entity}, which enable users to specify for example keywords to control information included in generated summaries, and help provide a better quality summary with fewer hallucinated entities. 

Unlike our approach, previous controllable summarization methods \cite{fan-etal-2018-controllable, he-etal-2022-ctrlsum} are often evaluated compared to generic summarization methods through qualitative analysis by human evaluators, which may suffer from biases. In this study, we combine qualitative and quantitative metrics of factual consistency at the entity level following \citet{nan-etal-2021-entity} and \citet{xiao2022entity}. We analyze the factual quality of summaries using controllable entity-centric methods compared with generic supervised methods and prompt-based methods.

\section{Annotation process}
Our dataset, \corpname{}, is based on the open access GUM \cite{zeldes2017gum}, a manually annotated multilayer corpus with Universal Dependencies (UD) parses \cite{de-marneffe-etal-2021-universal}, entity information (entity types, Wikification links and more), coreference resolution and discourse parses, as well as a human-written summary for each document. Data comes from 12 text types and covers over 200K tokens (see Table~\ref{tb:1}). 
\begin{table*}[htb]
\centering
\scalebox{0.9}{
\begin{tabular}{@{}lrrrrrr@{}}
\toprule
 & \textbf{Documents} & \textbf{Mentions} & \textbf{Entities} & \textbf{\% Salient Entities} & \textbf{Avg \# of Entities} & \textbf{Tokens} \\ \midrule
\textbf{academic (ac)} & 18 & 5,046 & 3,067 & 3.13 & 170 & 17,169 \\
\textbf{bio (bi)} & 20 & 5,768 & 3,326 & 5.11 & 166& 18,213 \\
\textbf{conversation (cn)} & 14 & 4,094 & 1,352 & 9.62 & 97 & 16,416 \\
\textbf{fiction (fc)} & 19 & 4,974 & 2,344 & 7.04 & 123 & 17,510 \\
\textbf{interview (it)} & 19 & 5,211 & 2,604 & 5.18 & 137 &18,190 \\
\textbf{news (nw)} & 23 & 4,720 & 2,544 & 11.08 & 111 & 16,145 \\
\textbf{reddit (rd)} & 18 & 4,543 & 2,302 & 4.13 & 128 & 16,364 \\
\textbf{speech (sp)} & 15 & 4,847 & 2,550 & 5.88 & 170 & 16,720 \\
\textbf{textbook (tx)} & 15 & 4,719 & 2,881 & 5.41 & 192 &16,693 \\
\textbf{vlog (vl)} & 15 & 4,498 & 1,629 & 11.42 & 109 & 16,864 \\
\textbf{voyage (vy)} & 18 & 4,471 & 2,952 & 7.79 & 164 & 16,514 \\
\textbf{whow (wh)} & 19 & 4,468 & 2,348 & 11.33 & 124 & 17,081 \\ \midrule
\textbf{Total} & 213 & 57,359 & 29,899 & 7.26 & 146 & 203,879 \\\bottomrule
\end{tabular}}
\caption{Overview of \corpname{}. \% salient entities = number of salient entities / total number of entities; Avg entities per summary = \# of entities / \# of documents in genre.}
\label{tb:1}
\vspace{-10pt}
\end{table*}

\corpname{} adds a layer of entity salience labels to all named and non-named entities in GUM, annotated by three trained experts (PhDs/PhD students in Computational Linguistics). The main goal is to annotate a subset of entities as salient if they are mentioned in the summaries, regardless of subjective judgments about their importance. Annotators are asked to look at the source document as well as the human-written summary first, and then make a binary judgment on every mention as to whether it is in both. In this way, annotators mark a judgment for every mention in both documents. This approach, which goes back to \citealt{dunietz2014new}, assumes that if something is salient it should appear in the summary, and conversely, if it appears in the summary, it must be salient, since summaries are meant to be as short and informative as possible. This assumption is mirrored in GUM's summary guidelines\footnote{\url{https://wiki.gucorpling.org/gum/summarization}}. We reason that while this approach could over-generate, it should have high recall, since it would be hard to summarize a document while omitting salient entities. Despite this, we note that our approach still flags only a fraction of entities as salient (around 7\%, see Table \ref{tb:1}). Using the gold standard manual coreference clusters, we ensure that all mentions of each cluster (=entity) are included as salient mentions, meaning annotations are consistent with the coreference layer in terms of entities.
\par


\begin{table}[h]
\centering
\resizebox{0.43\textwidth}{!}{%
\begin{tabular}{lrllr}
\hline
\textbf{} & \multicolumn{4}{l}{\textbf{Percentage/ Cohen's $\kappa$ agreement}} \\ \hline
\textbf{ac} & \multicolumn{2}{l}{0.9979/0.9983} & \textbf{rd} & 0.9928/0.9906 \\ \hline
\textbf{bi} & \multicolumn{2}{l}{0.9846/0.9840} & \textbf{sp} & 0.9913/0.9897 \\ \hline
\textbf{cn} & \multicolumn{2}{l}{0.9780/0.9666} & \textbf{tx} & 0.9942/0.9931 \\ \hline
\textbf{fc} & \multicolumn{2}{l}{0.9684/0.9621} & \textbf{vl} & 0.9993/0.9983 \\ \hline
\textbf{it} & \multicolumn{2}{l}{0.9860/0.9834} & \textbf{vy} & 0.9976/0.9967 \\ \hline
\textbf{nw} & \multicolumn{2}{l}{0.8955/0.8889} & \textbf{wh} & 0.9902/0.9869 \\ \hline
\textbf{Total} & \multicolumn{4}{c}{0.9813/0.9782} \\ \hline
\end{tabular}%
}
\caption{Genre-breakdown inter-annotator agreement on the \corpname{} test set at entity level.}
\label{tb:2}
\end{table}

We also double-annotated 21,770 tokens of the data corresponding to the 24 test documents of the UD release of GUM, containing 3,283 entities ($\approx$ 10\% of the data in Table~\ref{tb:1}) with binary SEE labels (salient vs. non-salient). To measure inter-annotator agreement (IAA), we compute raw percentage agreement and Cohen's $\kappa$ agreement at entity level for all 12 genres (if any mention of the entity is considered salient, the entity is considered salient). Since entity mention spans are given in GUM, our task only involves matching such spans to the summaries, and we achieve very high IAA across 12 genres (0.981 for raw agreement and 0.978 for Cohen's $\kappa$ agreement), with most of the texts achieving an agreement score over 0.9 (see the genre-breakdown IAA scores in Table~\ref{tb:2}). While this indicates very reliable results, annotators did disagree on some difficult cases:

\begin{itemize}
\item \textbf{Canonical mentions}: Some entities are mentioned in the summary differently than in the text, which sometimes makes it hard for annotators to locate the right salient mention in the text. This usually happens when the entity is a singleton (only being mentioned once in the document). For example, one summary mentions ``demographic information about the respondents'', which does not appear in the text. In this case, annotators flagged the mention ``Demographic variables'' in the text as salient, since it was judged to refer to the same thing in context. By contrast, another summary mentioned ``the history of the concept of atoms'', but the nearest mention in the text, ``early ideas in Atomic Theory'' was deemed not equivalent in its denotation. 
\item \textbf{Lack of explicit speaker information}: This type of issue occurred frequently in conversations, where no explicit speaker information is given in the text supplied to annotators, who needed to track who is speaking. For example, if a summary mentions ``Miles tells his friends about...'', then all interlocutors (Miles and his friends) should be marked as salient. However, the pronouns (\textit{I} and \textit{you}) in the conversation do not unambiguously indicate `who is talking to whom'. In this case, annotators were provided with gold speaker information from the dataset to help make the right decision.
\item \textbf{Non-nominal mentions}: According to our guidelines\footnote{\url{https://wiki.gucorpling.org/gum/salience}}, for an entity to be considered salient it must be (i) a markable mention in the source document, which include referential NPs and verbal markables (if they are coreferred to by an NP)\footnote{See entity annotation guidelines here: \url{https://wiki.gucorpling.org/gum/entities}} (ii) a verbal event in the summary coreferent with a nominal event in the source document. If the entity is mentioned in the source document as a non-nominal mention, then it is only considered salient if it is referred back to by a pronoun or noun. For example, in Figure \ref{fig:sal} the summary mentions ``dancing lewdly'', which corefers with a non-nominal mention ``the whole dance'' in the source document. In this case, ``dancing lewdly'' will not be marked as salient because ``the whole dance'' is not coreferred to by a pronoun in the document. 
\item \textbf{Aggregate and specific mentions}: When documents enumerated the members of an aggregate set mentioned in the summary but not the document, we decided to include the members as salient. For example, ``the remaining three shuttles'' are mentioned in one summary, while the document contains the three specific shuttles (`Space shuttle Endeavour', `Discovery', and `Enterprise'). These are thus all marked as salient. 

\end{itemize}
\section{Experimental setup}
In order to evaluate the usefulness of SEE annotations we apply our data to the task of automatic summarization and test 1) whether system summaries capture gold salient entities identified by humans, and 2) whether SEE information can improve summarization quality. We evaluate the following models:

\paragraph{BRIO}
BRIO \cite{liu-etal-2022-brio} is a recent SOTA abstractive summarization model, trained and fine-tuned on three newswire datasets: the CNN/Daily Mail dataset (CNN/DM, \citealt{hermannCNN}), XSum \cite{narayan-etal-2018-dont}, and the NYT dataset \cite{AB2/GZC6PL_2008}. It uses a novel training paradigm that introduces a contrastive learning component to estimate the probability of the generated summaries more accurately.\par 

We chose the pre-trained XSum BRIO model{\interfootnotelinepenalty=500 \footnote{\url{Yale-LILY/brio-xsum-cased} on Huggingface}, which most closely resembles the style of GUM's single sentence summaries (cf.~Figure \ref{fig:sal}). We test whether the summaries generated by the model are able to capture gold salient entities in \corpname{} using the UD test partition (see Table~\ref{tb:3}). We also include summary level scores on the full dataset in Table~\ref{tb:4} to see whether SEE information can enhance summarization quality. 

\paragraph{CTRLSum}
CTRLSum \cite{he-etal-2022-ctrlsum} is a summarization model used for generating abstractive summaries. It is considered a controllable summarization method because it produces summaries based on user input, which can specify entities of interest (in the form of keywords), summary length, and questions that the summary should answer. The system is a fine-tuned version of the $\mbox {BART} \textsubscript{LARGE}$ model \cite{lewis-etal-2020-bart} based on three training datasets: CNN/DM, arXiv scientific papers \cite{cohan-etal-2018-discourse}, and BIGPATENT (patent documents, \citealt{sharma-etal-2019-bigpatent}). 

CTRLSum is designed to separate test-time user control of summarization and the training process. During training, summaries are conditioned on the source document and automatically extracted keywords. At test time, a \textit{control function} is applied to map control aspects to keywords, while model parameters from training remain unchanged. Thus CTRLSum differs from other controllable summarization methods in not requiring separate models for each control aspect, generalizing to new keywords at test time. \par

We use the pre-trained CTRLSum model\footnote{\url{https://github.com/salesforce/ctrl-sum}} in three scenarios: \textsc{gold}, \textsc{pred} and \textsc{zero}. For \textsc{gold} we use the 3 most frequently mentioned gold salient entities in each document\footnote{The choice of using the top 3 salient entities rather than all salient entities is because the minimum count of salient entities (i.e.~several documents only have 3), and therefore it represents a reasonable prompt for the GPT model, which would otherwise potentially be asked to generate more salient entities than the document contains, leading to precision errors.} as ``keywords'' (all unique mentions of these entities are used, excluding pronouns); in \textsc{pred} we generate predicted salient entities using GPT-4 (\citealt{openai2023gpt4}), a generative LLM that achieves human-level performance on a range of benchmarks, using the prompt `\textit{Find the top 3 salient entities in the following document.}', and in \textsc{zero} we test without adding salient entities. 

\paragraph{GPT-4}
GPT-4 \cite{openai2023gpt4} is the latest version of Generative Pre-trained Transformers at the time of writing. Although training details for GPT-4 are not released (incl. model size, architecture, dataset, training method, etc.), we know from technical reports \cite{openai2023gpt4} that it was trained using masking and reinforcement learning from human feedback (RLHF). \par

For a more robust comparison between the fine-tuned models (BRIO and CTRLSum) and prompt-based models, we control the length of GPT-4 prompts using the following prompt: \textit{Summarize the following article in }N\textit{ sentences.} Since BRIO's XSum model produces one-sentence outputs and CTRLSum summaries are mostly 2-3 sentences, we can compare GPT-4 summaries with both systems using the sentence-count length prompt \textit{N}. In order to test whether adding gold or predicted entities to the model helps generate better summaries, we use the following prompt format: \textit{Summarize the following article in N sentences. In your summary, make sure to include the following words: <gold or predicted entity 1,2,3>}. 

\section{Evaluation}
\label{Sec:eval}
In this section, we evaluate model performance and the impact of SEE on two aspects of summarization: Section~\ref{sec:4.1} shows a manual evaluation of entity-level performance (are all salient entities included in summaries?) on the test set\footnote{The entity level evaluation was performed only on the test set because it needed to be carried out manually and separately for each of the 7 system outputs, making evaluation of the full dataset unfeasible.} (24 documents/~6k entity mentions, Table~\ref{tb:3}), and Section~\ref{sec:4.2} shows the overall summary quality on the entire dataset (213 documents/~30K entity mentions) based on automatic metrics. 

\begin{table*}[htb]
\centering
\scalebox{0.9}{
\begin{tabular}{@{}cccc || ccc ||ccc|| ccc@{}}
\toprule
\multirow{3}{*}{} & $P_t$ & $R_t$ & $F1_t$ & $P_t$ & $R_t$ & $F1_t$ & $P_t$ & $R_t$ & $F1_t$ & $P_t$ & $R_t$ & $F1_t$\\
& \multicolumn{3}{c||}{\textbf{{CTRLGold}}} & \multicolumn{3}{c||}{\textbf{{CTRLPred}}} & \multicolumn{3}{c||}{\textbf{{CTRL0}}}& \multicolumn{3}{c}{\textbf{{BRIO}}} \\
\midrule
\textbf{ac} & 0.615 & 0.615 & 0.615 & 0.536 & \textcolor{blue}{0.583} & 0.558 & 0.857 & \textcolor{blue}{0.462} & \textcolor{blue}{0.600} & 0.571 & 0.333 & 0.421  \\
\textbf{bi} & 0.813 & \textcolor{blue}{0.765} & \textcolor{blue}{0.788} & 0.607 & 0.403 & 0.434 & 0.600 & 0.176 & 0.273 & 0.600 & 0.462 & 0.522 \\
\textbf{cn} & 0.471 & 0.500 & 0.485 & 0.583 & 0.339 & 0.413 & 0.500 & 0.250 & 0.333 & 0.330 & 0.250 & 0.284 \\
\textbf{fc} & 0.583 & 0.500 & 0.538 & 0.750 & 0.417 & 0.533 & \textcolor{red}{0.333} & 0.214 & 0.261 & \textcolor{red}{0.166} & 0.154 & 0.160 \\
\textbf{it} & 0.769 & 0.526 & 0.625 & \textcolor{blue}{0.900} & 0.436 & \textcolor{blue}{0.564} & 0.875 & 0.368 & 0.519 & 0.647 & \textcolor{blue}{0.478} & \textcolor{blue}{0.550}\\
\textbf{nw} & 0.632 & 0.500 & 0.558 & 0.833 & 0.350 & 0.452 & 0.900 & 0.375 & 0.529 & 0.636 & 0.292 & 0.400 \\
\textbf{rd} & 0.500 & 0.643 & 0.563 & 0.875 & 0.354 & 0.500 & 0.600 & 0.214 & 0.316  & 0.455 & 0.385 & 0.417 \\
\textbf{sp} & \textcolor{red}{0.217} & 0.481 & \textcolor{red}{0.299} & 0.486 & 0.217 & 0.299 & 0.857 & 0.222 & 0.353 & \textcolor{blue}{0.666} & 0.296 & 0.410 \\
\textbf{tx} & 0.632 & 0.750 & 0.686 & 0.500 & 0.198 & 0.283 & \textcolor{red}{0.333} & 0.125 & 0.182 & 0.222 & 0.133 & 0.166\\
\textbf{vl} & 0.800 & 0.615 & 0.696 & 0.833 & 0.244 & 0.377 & 0.667 & 0.154 & 0.250 & 0.538 & 0.292 & 0.379\\
\textbf{vy} & 0.577 & \textcolor{red}{0.455} & 0.508 & \textcolor{red}{0.479} & 0.292 & 0.360 & 0.500 & 0.094 & 0.158 & 0.200 & \textcolor{red}{0.066} & \textcolor{red}{0.099}\\
\textbf{wh} & \textcolor{blue}{0.857} & 0.514 & 0.643 & 0.500 & \textcolor{red}{0.153} & \textcolor{red}{0.229} & \textcolor{blue}{1.000} & \textcolor{red}{0.057} & \textcolor{red}{0.108} & 0.500 & 0.152 & 0.233\\ 
\midrule 
\textbf{Total} & 0.555 & 0.555 & 0.555 & 0.657 & 0.332 & 0.417 & 0.658 & 0.206 & 0.313 & 0.512 & 0.255 & 0.340\\ 
\hline 
& \multicolumn{3}{c||}{\textbf{{GPTGold}}} & \multicolumn{3}{c||}{\textbf{{GPTPred}}} & \multicolumn{3}{c||}{\textbf{{GPT0}}} \\
\cmidrule{1-10}
\textbf{ac}& \textcolor{red}{0.409} & {0.692} & {0.514} & {0.400} & {0.615} & {0.485} & \textcolor{red}{0.304} & {0.538} & {0.389} \\
\textbf{bi}& {0.619} & {0.765} & {0.684} & {0.375} & {0.529} & {0.439} & {0.455} & {0.588} & {0.513} \\
\textbf{cn}&{0.524} & {0.688} & {0.595} & {0.435} & {0.625} & {0.513} & {0.333} & {0.438} & {0.378} \\
\textbf{fc}&{0.650} & \textcolor{blue}{0.929} & \textcolor{blue}{0.765} & {0.391} & {0.643} & {0.486} & {0.409} & {0.643} & {0.500} \\
\textbf{it} &\textcolor{blue}{0.737} & {0.737} & {0.737} & {0.364} & {0.632} & {0.462} & {0.400} & {0.526} & {0.455} \\
\textbf{nw} & {0.647} & {0.458} & {0.537} & {0.462} & {0.500} & {0.480} & {0.481} & {0.542} & {0.510} \\
\textbf{rd} & {0.650} & \textcolor{blue}{0.929} & \textcolor{blue}{0.765} & {0.579} & \textcolor{blue}{0.786} & \textcolor{blue}{0.667} & {0.647} & \textcolor{blue}{0.786} & \textcolor{blue}{0.710} \\
\textbf{sp} &{0.459} & {0.630} & {0.531} & \textcolor{red}{0.300} & \textcolor{red}{0.222} & \textcolor{red}{0.255} & {0.432} & {0.593} & {0.500} \\
\textbf{tx}&{0.571} & {0.750} & {0.649} & {0.367} & {0.688} & {0.478} & {0.423} & {0.688} & {0.524} \\
\textbf{vl}&{0.450} & \textcolor{red}{0.346} & \textcolor{red}{0.391} & {0.550} & {0.423} & {0.478} & {0.421} & {\textcolor{red}{0.308}} & \textcolor{red}{0.356} \\
\textbf{vy}&{0.433} & {0.394} & {0.413} & {0.533} & {0.485} & {0.508} & {0.615} & {0.485} & {0.542} \\
\textbf{wh}&{0.696} & {0.457} & {0.552} & \textcolor{blue}{0.690} & {0.571} & {0.625} & \textcolor{blue}{0.760} & {0.543} & {0.633} \\ 
\cmidrule{1-10}
\textbf{Total} & 0.570 & 0.648 & \textbf{0.594} & 0.454 & 0.560 & 0.490 & 0.473 & 0.556 & 0.501 \\
\cmidrule{1-10}
\end{tabular}}
\caption{Entity level scores and the macro-averaged scores per model on the \corpname{} test set for several systems. The \textcolor{blue} {blue} text is the highest score across 12 genres and \textcolor{red} {red} text is the lowest. The top $F1_t$ score across all models is bolded. See Table~\ref{tb:1} for genre codes.}
\label{tb:3}
\end{table*}

\subsection{Entity Level Evaluation} \label{sec:4.1}
We use \corpname{} to test the two systems above, as well as GPT-4 itself, 
and examine whether baseline results differ from settings where predicted or gold-standard salient entities are provided (for CTRLSum and GPT-4; BRIO does not provide summary control mechanisms). Apart from investigating whether the systems are able to capture entities that appear in the summary (see Table~\ref{tb:3}), we conducted a quantitative (see Appendix~\ref{sec: quant-hallucination} for additional quantitative analysis of system output factuality using automated scores i.e.~$SummaC_{Conv}$ \cite{laban2022summac}) and qualitative analysis on entity hallucination (see Figures \ref{fig:exfic},\ref{fig:exconvall},\ref{fig:exTx}), which examine entities that didn't appear in the summary or source document.\par

Following \citet{nan-etal-2021-entity}, we evaluate summaries at the entity level by taking  precision, recall and F1 score for unique predicted entities (rather than mentions), e.g.~$P_{t} = N(h \cap t)/N(h)$, is the precision, where 
$N(h \cap t)$ is the number of distinct gold salient entities also mentioned in the summary and 
$N(h)$ is the number of entities mentioned in the generated summary. For all the system summaries, we performed a manual evaluation to ensure the quality of mention/entity detection in all 12 genres. That is, the mentions/entities in the generated summaries are identified manually by one of the authors rather than automatically by an entity resolver or coreference system, which are known to perform poorly on out-of-domain genres \cite{moosavi-strube-2017-lexical, zhu2021ontogum}. \par

Overall, we see that both dedicated summarization systems and prompt-based LLMs show poor performance in capturing all salient entities, with $F1$ scores ranging from 30s to 50s. Table~\ref{tb:3} shows that BRIO trained on XSum \cite{narayan-etal-2018-dont} performs poorly in all 12 genres, but especially in genres rich in conversations (\textit{conversation} and \textit{fiction}).\footnote{Although \textit{fiction} is considered a written genre, the data contains substantial dialogue between characters.} This is expected, as models trained solely on news may not generalize to out-of-domain (OOD) data like \textit{conversation} and \textit{fiction}. Interestingly, we also found that entity hallucination is most severe in these genres, see e.g.~$P$ score of 0.166 for \textit{fiction}, mainly due to hallucinations. For example, the BRIO summary in Figure~\ref{fig:exfic} for one of the \textit{fiction} mentions `the German writer and photographer Barbara Hepworth' even though it has not been appeared in the source document. \par 

We also tested whether providing salient entities to the model would improve performance. Table~\ref{tb:3} shows CTRLSum and GPT scores in three settings: adding gold salient entities that have the top 3 frequent mentions in the document (\textsc{gold}), adding predicted salient entities from the GPT-4 model (\textsc{pred}), and no salient entities provided (\textsc{zero}). Unsurprisingly, GPT with gold salient entities (GPTGold) outperforms all models, with $F1=0.594$. The models in the \textsc{gold} setting also outperform those with the other two settings, as can be seen in the $F1$ scores of CTRLSum methods and GPT methods. Interestingly, despite having a lower $F1$ score, CTRL0 has a surprisingly high $P$ score, while CTRLGold has the lowest $P$ score. This is because CTRL0 often picks out the first sentence in the document as the generated summary, which usually contains a large number of salient entities (high precision) but not all of the important ones (low recall), as shown in Figure \ref{fig:exconvall}. We do not see this pattern in GPT methods, suggesting that the position of entities is not the only factor to take into account for GPT-4 to generate summaries.\par 

For genre comparison, most of the models perform relatively well in written genres (e.g.~\textit{academic, biography, interview\footnote{Although \textit{interview} is considered a spoken genre, the source of the data is Wikinews interviews with politicians, which makes the language similar to news articles.}}) but not in spoken genres (e.g.~\textit{speech}). This is reasonable, as spoken genres are considered ``unfamiliar'' and out-of-domain for models trained on written data. However, this modality (written vs. spoken) effect does not seem to explain the performance of GPT methods, as can be seen in the lower scores of both spoken (\textit{speech}, \textit{vlog}) and written (\textit{academic}) genres. The poor performance of \textit{academic} and \textit{speech} might be explained by the fact that they have a rather low \% of salient entities and a rather large number of entities per document (see Table \ref{tb:1}), which makes it hard for the model to capture the targeted salient entities in a document. Surprisingly, GPT methods perform well in \textit{fiction} and \textit{reddit}, which are usually hard for other models. This might be because the Pile dataset \cite{DBLP:journals/corr/abs-2101-00027}, which is known to be used as training data for GPT models, includes diverse data sources including books and web text.\par

We also saw that adding gold salient entities to CTRLSum methods is especially beneficial for genres like \textit{voyage} and \textit{wikihow}. However, we did not see the power of adding gold salient entities for \textit{voyage} and \textit{wikihow} in GPT methods. Without adding gold entities, CTRL0 and CTRLPred models often produce summaries that are too short and abstractive, leaving out important details. By contrast, GPT summaries in all three settings are rather similar in terms of the mentioned entities. For example, the CTRL0 summary for one of the \textit{wikihow} documents is simply its title: ``\textit{How to Grow Beavertail Cactus.}'', while the CTRLGold and GPT summaries mention methods and materials that can be used to grow Beavertail Cactus, which human summarization also captured.\par

Interestingly, we observe that adding gold salient entities to GPT models is specifically useful for highly conversational genres like \textit{conversation} and \textit{interview}, whereas predicted entities added to the model in these genres are not as useful as the gold ones. This suggests that predicting such entities is still a difficult task for GPT-4, especially in spoken genres. A closer look at these predicted entities shows that GPT-4 tends to pick out \textsc{person} entities in the document as salient, which is not always correct. Figure~\ref{fig:exconvall} shows the predicted entities (in italics) including several \textsc{person} entities, which were disregarded by humans. \par
In terms of entity hallucination, we can see from Figures \ref{fig:exfic},\ref{fig:exconvall},\ref{fig:exTx} that BRIO summaries contain the most hallucinated entities, while we hardly see any hallucinations in CTRLSum and GPT-4 summaries. Our quantitative analysis in Appendix~\ref{sec: quant-hallucination} also shows that adding salient entities to the model enhances the faithfulness of the summaries. However, it is worth noting that our analysis focuses on `intrinsic' hallucinations \cite{ji2023survey}, which are those that do not appear in and/or contradict the source document. We did notice other types of hallucinations (i.e.~`extrinsic' ones) in GPT-4 outputs. These include entities neither supported nor contradicted by the source document. For example, in Figure~\ref{fig:exconvall}, GPTPred outputs `the speaker's ``long'' friendship with…,' although the document does not specify whether the friendship is ``long'' or not. We believe that further work could be done on central propositions or claims in text and their role in curbing this type of hallucination but a rigorous evaluation of this issue lies beyond the scope of the experiments we conducted. 

\subsection{Summary Level Evaluation}
\label{sec:4.2}

\begin{table*}[h]
\centering
\resizebox{\textwidth}{!}{%
\begin{tabular}{clrrrrrrrrrrrrr}
\hline
\textbf{Model} & \textbf{Metrics} & \textbf{ac} & \textbf{bi} & \textbf{cn} & \textbf{fc} & \textbf{it} & \textbf{nw} & \textbf{rd} & \textbf{sp} & \textbf{tx} & \textbf{vl} & \textbf{vy} & \textbf{wh} & \textbf{Avg} \\ \hline
\multirow{3}{*}{\textbf{BRIO}} & \textbf{ROUGE-1} & 31.49 & 30.44 & 14.81 & \textcolor{red}{11.77} & \textcolor{blue}{31.53} & 30.66 & 19.54 & 27.25 & 14.69 & 15.56 & 18.89 & 15.60 & 21.85 \\
{} & {\textbf{ROUGE-2}} & {10.06} & {12.96} & {2.27} & {1.94} & {\textcolor{blue}{13.43}} & {11.79} & {2.89} & {11.39} & {\textcolor{red}{1.57}} & {5.48} & {3.63} & {4.59} & {6.83} \\
 & \textbf{ROUGE-L} & 23.94 & 25.11 & 11.49 & \textcolor{red}{8.97} & \textcolor{blue}{26.59} & 23.70 & 17.27 & 23.42 & 11.69 & 13.32 & 15.51 & 13.70 & 17.89 \\
 & \textbf{BERTScore} & .65 & .65 & .55 & \textcolor{red}{.51} & \textcolor{blue}{.66} & .63 & .56 & .61 & .56 & .56 & .58 & .54 & .59 \\
\hdashline\multirow{3}{*}{\textbf{CTRL0}} & \textbf{ROUGE-1} & 25.11 & 32.92 & \textcolor{red}{7.18} & 8.66 & \textcolor{blue}{35.02} & 29.22 & 12.50 & 15.41 & 15.29 & 10.13 & 17.39 & 16.89 & 18.81 \\
{} & {\textbf{ROUGE-2}} & {5.32} & {17.30} & {2.72} & {1.46} & {\textcolor{blue}{18.73}} & {11.08} & {1.56} & {8.49} & {4.49} & {\textcolor{red}{0.51}} & {5.51} & {6.25} & {6.95} \\
 & \textbf{ROUGE-L} & 20.85 & \textcolor{blue}{31.98} & \textcolor{red}{6.16} & 7.94 & 29.38 & 23.70 & 11.70 & 13.66 & 13.81 & 7.47 & 15.54 & 15.80 & 16.50 \\
 & \textbf{BERTScore} & .59 & \textcolor{blue}{.65} & \textcolor{red}{.45} & .49 & .64 & .63 & .51 & .54 & .52 & .50 & .56 & .53 & .55 \\
\hdashline\multirow{3}{*}{\textbf{CTRLPred}} & \textbf{ROUGE-1} & 23.20 & \textcolor{blue}{38.14} & \textcolor{red}{16.61} & 23.49 & 34.65 & 31.58 & 20.85 & 21.89 & 21.31 & 17.92 & 23.57 & 18.76 & 24.33 \\
{} & {\textbf{ROUGE-2}} & {8.44} & {\textcolor{blue}{20.31}} & {\textcolor{red}{4.36}} & {4.59} & {18.98} & {13.90} & {5.43} & {9.34} & {6.30} & {7.38} & {6.58} & {4.87} & {9.21} \\
 & \textbf{ROUGE-L} & 19.27 & \textcolor{blue}{34.31} & \textcolor{red}{13.86} & 20.35 & 31.20 & 25.58 & 18.72 & 18.24 & 18.38 & 15.33 & 19.78 & 16.46 & 20.96 \\
 & \textbf{BERTScore} & .55 & \textcolor{blue}{.67} & \textcolor{red}{.47} & .51 & .63 & .61 & .50 & .55 & .51 & .51 & .56 & .51 & .55 \\
\hdashline\multirow{3}{*}{\textbf{CTRLGold}} & \textbf{ROUGE-1} & 29.48 & \textcolor{blue}{42.44} & 20.05 & \textcolor{red}{19.42} & 39.76 & 41.07 & 23.60 & 27.51 & 23.58 & 24.35 & 28.26 & 26.12 & 28.80 \\
{} & {\textbf{ROUGE-2}} & {7.83} & {\textcolor{blue}{25.67}} & {5.62} & {\textcolor{red}{2.66}} & {18.24} & {18.32} & {4.60} & {13.10} & {7.09} & {6.26} & {12.12} & {6.83} & {10.69} \\
 & \textbf{ROUGE-L} & 22.72 & \textcolor{blue}{39.91} & 18.14 & \textcolor{red}{17.40} & 31.67 & 31.06 & 18.10 & 23.96 & 19.32 & 17.75 & 23.31 & 21.86 & 23.77 \\
 & \textbf{BERTScore} & .59 & \textcolor{blue}{.69} & \textcolor{red}{.50} & .52 & .65 & .68 & .54 & .57 & .57 & .53 & .60 & .56 & .58 \\
\hdashline\multirow{3}{*}{\textbf{GPT0 N=1}} & \textbf{ROUGE-1} & 29.89 & 46.20 & \textcolor{red}{18.79} & 29.67 & 38.04 & \textcolor{blue}{46.69} & 28.97 & 32.26 & 28.81 & 25.30 & 29.27 & 35.90 & 32.48 \\
{} & {\textbf{ROUGE-2}} & {9.72} & {25.19} & {\textcolor{red}{4.23}} & {5.37} & {15.21} & {\textcolor{blue}{20.99}} & {6.05} & {15.35} & {9.13} & {8.99} & {7.64} & {10.35} & {11.52} \\
 & \textbf{ROUGE-L} & 22.86 & 37.97 & \textcolor{red}{16.91} & 23.05 & 28.12 & \textcolor{blue}{36.07} & 22.79 & 25.62 & 23.09 & 20.75 & 22.27 & 28.68 & 25.68 \\
 & \textbf{BERTScore} & .64 & .72 & \textcolor{red}{.59} & .64 & .69 & \textcolor{blue}{.73} & .64 & .66 & .64 & .64 & .66 & .67 & .66 \\
\hdashline\multirow{3}{*}{\textbf{GPTPred N=1}} & \textbf{ROUGE-1} & 29.49 & 41.35 & \textcolor{red}{21.27} & 30.46 & 38.56 & \textcolor{blue}{46.18} & 31.04 & 29.78 & 24.86 & 40.84 & 31.83 & 33.12 & 33.23 \\
{} & {\textbf{ROUGE-2}} & {9.83} & {20.28} & {\textcolor{red}{4.29}} & {9.07} & {14.51} & {\textcolor{blue}{20.84}} & {6.25} & {10.00} & {5.75} & {15.96} & {9.71} & {8.86} & {11.28} \\
 & \textbf{ROUGE-L} & 24.29 & \textcolor{blue}{33.77} & 18.30 & 24.97 & 28.42 & 32.81 & 24.48 & 22.76 & \textcolor{red}{17.94} & 31.29 & 24.52 & 27.58 & 25.93 \\
 & \textbf{BERTScore} & .64 & .70 & .62 & .65 & .69 & \textcolor{blue}{.72} & .63 & .65 & \textcolor{red}{.61} & .69 & .66 & .67 & .66 \\
\hdashline\multirow{3}{*}{\textbf{GPTGold N=1}} & \textbf{ROUGE-1} & 33.96 & \textcolor{blue}{47.63} & \textcolor{red}{24.05} & 30.57 & 39.00 & 47.17 & 30.78 & 30.55 & 28.59 & 31.01 & 33.10 & 34.30 & \textbf{34.23} \\
{} & {\textbf{ROUGE-2}} & {10.96} & {\textcolor{blue}{27.00}} & {5.84} & {\textcolor{red}{5.62}} & {16.04} & {20.46} & {7.92} & {12.96} & {9.21} & {11.78} & {12.47} & {10.30} & {\textbf{12.55}} \\
 & \textbf{ROUGE-L} & 24.78 & \textcolor{blue}{40.68} & 21.81 & 23.39 & 30.54 & 33.94 & 23.72 & 24.15 & \textcolor{red}{20.46} & 25.15 & 26.86 & 26.03 & \textbf{26.79} \\
 & \textbf{BERTScore} & .66 & \textcolor{blue}{.73} & \textcolor{blue}{.62} & .65 & .70 & .72 & .64 & .69 & .63 & .64 & .68 & .67 & \textbf{.67} \\ \hline
\end{tabular}
}
\caption{Summary level scores on \corpname{} with the SOTA abstractive summarization method (BRIO), controllable summarization method (CTRLSum), and zero-shot LLM GPT-4 with three different settings: with gold salient entity information (\textsc{gold}), with predicted salient entities from GPT-4 (\textsc{pred}) and without salient entity information (\textsc{zero}). N represents the sentence-count length in GPT-4 methods. The \textcolor{blue} {blue} text is the highest score across 12 genres and \textcolor{red} {red} text is the lowest. The highest average scores across all models are bolded.}
\label{tb:4}
\end{table*}
We evaluate the quality of the generated summaries from SOTA models and prompt-based models, including BRIO, CTRLSum and GPT-4, using the widely used ROUGE scores \cite{lin-2004-rouge} and BERTScore \cite{DBLP:conf/iclr/ZhangKWWA20}. ROUGE-1 and ROUGE-2 are used to measure the unigram and bigram overlap with the reference summary, respectively. ROUGE-L score (longest common subsequence) is used to measure the sentence level structural similarity between the generated and reference summaries. BERTScore measures the semantic similarity between the generated and reference summaries by computing the similarity score for each token in the generated and reference summary. \par

In general, we found that all models perform the best with the \textsc{gold} setting, and the \textsc{zero} setting has the lowest performance. This is unsurprising, as adding gold salient entities to the model enhances both lexical/content overlap and semantic similarity between the generated summary and the ground truth summary. \par

As can be seen in Table~\ref{tb:4}, GPTGold outperforms all the other models on all metrics. Within GPT methods, we found that models prompted to summarize in 1 sentence generally outperform those prompted in 2-3 sentences (compare the numbers in Table~\ref{tb:4} with those in Table~\ref{tb:c23}). This is because longer summaries are usually too specific, leading to lower quality summaries. Figure \ref{fig:exconvall} shows a qualitative example of this. \par

In terms of the differences between models, we observe that CTRLSum summaries are more extractive than BRIO summaries, containing more document entities (predicted or gold) in the output. We also found that BRIO summaries suffer a lot from entity hallucination, which can be alleviated by CTRLSum methods. GPT-4 summaries are considered the closest to the gold summaries for the following reasons: First, they contain as many gold entities as the CTRLSum summaries but with more subjective coherence. Second, they have the least hallucinated entities compared to the other two models. See Figure \ref{fig:exconvall} for a comparison between the ground truth summary and summaries generated by all the models from a \textit{conversation} document in our dataset. \par
\begin{figure*}[h!]
\centering
\includegraphics[width=1\textwidth, height=90mm] {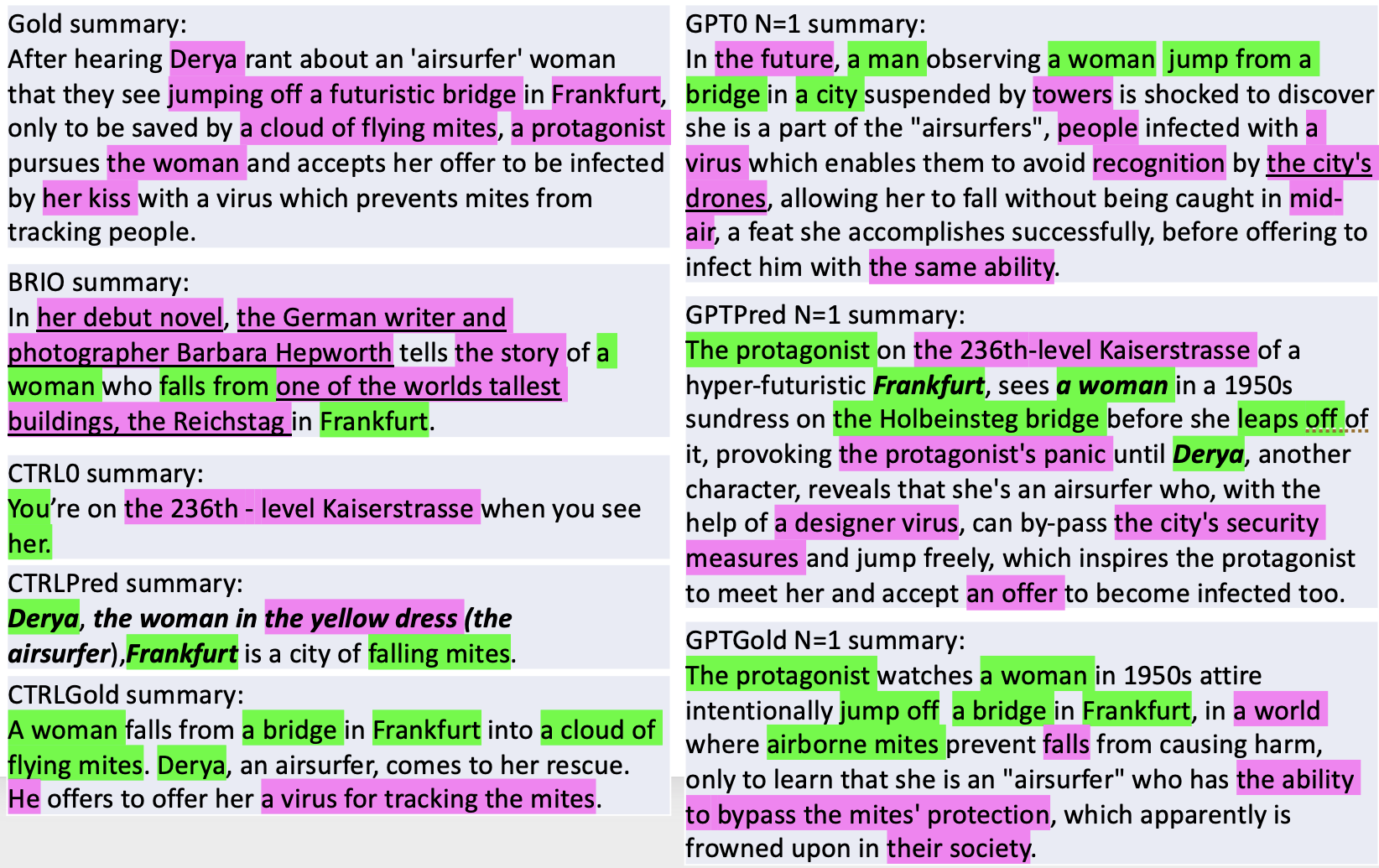}
\caption{Generated summaries with several models from a \textit{fiction} document in \corpname{}. The first mention of the entity has been highlighted in pink. Those that have a match in the ground truth summary are highlighted in green. The hallucinated entities are underlined and predicted entities by GPT-4 are in italics.}
\label{fig:exfic}
\end{figure*}
For genre comparison, all the models perform relatively well in written genres such as \textit{biography}, \textit{interview}, and \textit{news} since they are trained on written genre data. Although the training details for GPT-4 are unknown, it is likely that the majority of data it has been trained on is relatively similar to news language. For example, we see that GPT0 and GPTPred methods perform extremely well on \textit{news} data (see the blue text in news genre in Table \ref{tb:4}). However, we see that most of the models perform poorly on \textit{conversation} and \textit{fiction} data, which shows that both the SOTA models and prompt-based models are less familiar with spoken genres and are thus incapable of generating high-quality summaries for these ``outlier'' genres. Specifically, we found that BRIO summaries for \textit{fiction} data suffer from severe entity hallucination problems, which makes the generated summary not factually consistent with the gold one. GPT summaries, on the other hand, suffer less from entity hallucinations but provide overly specific details of document contents (e.g.~\textit{the protagonist's panic}, \textit{the city's security measures} in Figure \ref{fig:exfic}). Similar to GPT summaries, CTRLSum summaries have few hallucinated entities. However, they are usually too short to cover a more exhaustive overview of document content.
Figure \ref{fig:exfic} shows the ground truth summary for one of the \textit{fiction} documents with generated summaries from all the models. The hallucinated entities are underlined and predicted entities from GPT-4 are in italics.

\section{Conclusion}
\label{sec:5}
This paper presented \corpname{}, the first manually annotated entity salience dataset covering all named and non-named entities for 12 genres of English text. \corpname{} achieves a high level of agreement in all 12 genres, creating a high-quality entity salience dataset that allows the evaluation of SEE annotations in diverse genres. Our evaluation shows that a significant amount of salient entities are not captured by SOTA abstractive summarization models and prompt-based LLMs and that adding salient entities to model inputs substantially enhances the coverage. We also show that adding such entities helps reduce hallucinations in less common genres (e.g.~textbooks and travel guides) to a large extent, generating higher-quality summaries. We hope that \corpname{} will enable further research on entity salience and can serve as a challenging dataset for testing text summarization methods in a wide range of genres focusing on entities.

\section*{Limitations}

This paper has several limitations. First and most important is the restriction of the data to English, the highest resource language in NLP research -- it is likely that our findings underestimate the contribution of providing salient entities for summarization in lower resource languages, while also overestimating the performance of pretraind models on the summarization and salient entity prediction tasks for the same languages. It is also possible that SEE annotation would not generalize well, or suffer from more disagreements in other languages, though we believe this is unlikely. 

A further limitation in terms of the evaluation of pretrained LLMs is that we cannot rule out that models have seen some of the evaluation data in some form during pretraining. GUM data is part of the Universal Dependencies project, which is managed over GitHub, and is therefore susceptible to inclusion in the Pile dataset, known to be used as training data for GPT models. If such effects are present in our evaluation, they should minimize, rather than maximize the contribution of providing SEE information. Our data is also relatively small in terms of summarization datasets, meaning that while it may not substantially affect LLM training, more data would lead to better results.

Additionally, we would like to point out that the summary level evaluation in Section~\ref{sec:4.2} could benefit from a human evaluation study on the quality of the system and reference summaries. First, it has been pointed out in \citet{goyal2022news} that automatic metrics (e.g.~ROUGE, BERTScore), though being commonly used, may not always correlate well with human evaluation. In this respect, we conducted the first manual, qualitative evaluation of the system summaries for the included entities (see Figures \ref{fig:exfic},\ref{fig:exconvall},\ref{fig:exTx}). Our analysis shows that adding predicted or gold salient entities to summarization models helps enhance summary quality by alleviating hallucinations in summaries. Despite this, we certainly believe that a more systematic human evaluation of system summaries would be beneficial, and it's worth exploring in future work. Second, previous research \cite{liu2022revisiting, pu2023summarization} has found that the reference summaries in the existing datasets are not always of good quality, especially when compared with summaries generated by LLMs. The expert-written summaries in GUM (see \citet{LiuZeldes2023} for more details), however, are considered high quality because the summaries follow stricter guidelines than other `found' summarization datasets \cite{gamon2013identifying}. This is supported by the human evaluation study conducted in \citet{LiuZeldes2023}, where human evaluators strongly preferred GUM human-written summaries to summaries generated by LLMs such as GPT-3. Also, GUM summaries were found to be best at substituting reading the text, while summaries from LLMs and pre-trained supervised models were considered less substitutive.

Finally, we note that the reference-based summarization paradigm is fundamentally limited in scoring outputs based on gold standard comparisons, despite the fact that alternative summaries may be equally good. We counter this issue by performing manual, qualitative human evaluation in this paper, and argue that while different summaries may include other ancillary entities, ones that are truly salient are likely to appear in almost any valid summary of a document, suggesting that at least the SEE recall of our manual approach should be satisfactory. We feel that this is valuable new data that can contribute especially to existing, automatically constructed datasets using click data \cite{gamon2013identifying}, NER/coreference resolution tools \cite{dunietz2014new} or hyperlinks \cite{wu2020wn}, which have also covered rather few domains in the past, and no spoken data. We leave the study of precision in SEE with multiple reference summaries for future papers.

\bibliography{anthology,custom}
\bibliographystyle{acl_natbib}

\appendix

\section{GPT-4 summary level scores with different length constraints}
\label{sec:appendix-full-data}
We found that GPT models prompted to summarize in 1 sentence (N=1) usually outperform those prompted in 2-3 sentences (N=2,3). Compare Table~\ref{tb:c23} with the GPT results in Table \ref{tb:4}. Figure~\ref{fig:exconvall} shows qualitative differences between GPT-4 models with different length controls.

\begin{table*}[h]
\centering
\resizebox{\textwidth}{!}{%
\begin{tabular}{clrrrrrrrrrrrrr} 
\hline
\textbf{Model} & \textbf{Metrics} & \textbf{ac} & \textbf{bi} & \textbf{cn} & \textbf{fc} & \textbf{it} & \textbf{nw} & \textbf{rd} & \textbf{sp} & \textbf{tx} & \textbf{vl} & \textbf{vy} & \textbf{wh} & \textbf{Avg} \\ 
\hline
\multirow{3}{*}{\textbf{GPT0 N=2,3}} & \textbf{ROUGE-1} & 28.58 & 43.39 & \textcolor{red}{22.51} & 27.88 & 34.23 & \textcolor{blue}{43.61} & 26.67 & 31.27 & 29.34 & 23.73 & 29.30 & 37.66 & 31.51 \\
& \textbf{ROUGE-2} & 8.35 & \textcolor{blue}{25.17} & 5.06 & 5.24 & 13.88 & 18.12  & \textcolor{red}{4.13} & 15.45 & 8.49 & 9.20 &  7.93 & 11.57 & 11.05\\
 & \textbf{ROUGE-L} & 22.85 & \textcolor{blue}{36.13} & 20.26 & 21.89 & 26.20 & 33.00 & 20.59 & 23.99 & 22.43 & \textcolor{red}{19.91} & 23.29 & 29.27 & 24.98 \\
 & \textbf{BERTScore} & .62 & \textcolor{blue}{.69} & .60 & .63 & .64 & .69 & .60 & .63 & \textcolor{red}{.59} & .61 & .65 & .66 & .63 \\ 
\hdashline
\multirow{3}{*}{\textbf{GPTPred N=2,3}} & \textbf{ROUGE-1} & 29.61 & 43.21 & \textcolor{red}{22.47} & 26.73 & 37.44 & \textcolor{blue}{44.15} & 27.92 & 28.33 & 23.46 & 24.36 & 27.54 & 34.03 & 30.77 \\
& \textbf{ROUGE-2} & 9.98 & \textcolor{blue}{21.76} & \textcolor{red}{3.75} & 5.32 & 15.76 & 18.17 & 5.01 & 12.35 & 5.09 & 8.97 & 7.76 & 10.37 & 10.36 \\
 & \textbf{ROUGE-L} & 22.95 & \textcolor{blue}{35.41} & 19.07 & 20.42 & 27.74 & 32.39 & 22.04 & 23.16 & \textcolor{red}{17.13} & 18.77 & 22.27 & 28.27 & 24.14 \\
 & \textbf{BERTScore} & .61 & \textcolor{blue}{.68} & .59 & .61 & .65 & .68 & .60 & .63 & \textcolor{red}{.57} & .62 & .64 & .65 & .63 \\ 
\hdashline
\multirow{3}{*}{\textbf{GPTGold N=2,3}} & \textbf{ROUGE-1} & 32.01 & 43.29 & \textcolor{red}{23.87} & 28.58 & 37.72 & \textcolor{blue}{46.31} & 29.00 & 29.67 & 28.22 & 26.44 & 29.92 & 36.19 & 32.60 \\
& \textbf{ROUGE-2} & 11.25 & \textcolor{blue}{23.80} & 4.64 & 5.04 & 12.75 & 19.95 & \textcolor{red}{4.24} & 13.91 & 7.13 & 10.38 & 11.64 & 11.19 & 11.33 \\
 & \textbf{ROUGE-L} & 23.69 & \textcolor{blue}{34.90} & 20.79 & 20.01 & 27.65 & 32.39 & 21.81 & 24.27 & \textcolor{red}{19.53} & 21.09 & 24.99 & 28.38 & 24.96 \\
 & \textbf{BERTScore} & .62 & .69 & \textcolor{red}{.57} & .62 & .65 & \textcolor{blue}{.70} & .61 & .63 & .59 & .63 & .65 & .66 & .64 \\\hline
\end{tabular}}
\caption{Summary level scores on \corpname{} with GPT-4 N=2,3. The \textcolor{blue} {blue} text is the highest score across 12 genres and \textcolor{red} {red} text is the lowest.}
\label{tb:c23}
\end{table*}

\begin{figure*}[h]
\centering
\includegraphics[width=\textwidth, height=50mm]{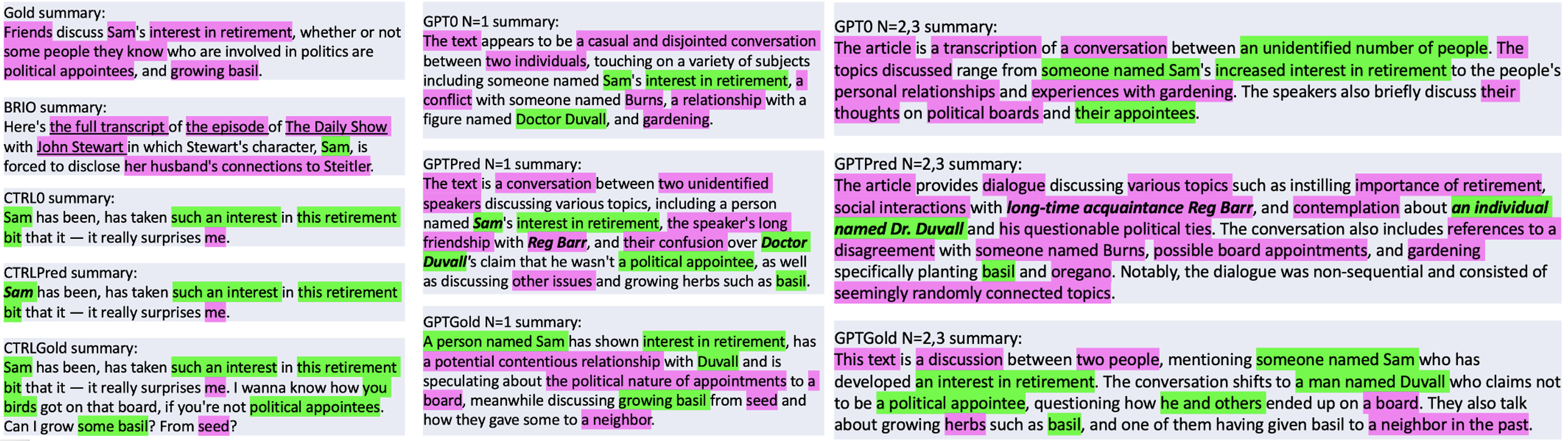}
\caption{An example of generated summaries with all the models from a \textit{conversation} document in \corpname{}. The first mention of the entity has been highlighted in pink. Those that have a match in the ground truth summary are highlighted in green. The hallucinated entities are underlined and predicted entities by GPT-4 are in italics.} \label{fig:exconvall}
\end{figure*}

\section{API costs}
At the time of running our experiments, GPT-4 API costs \$0.06 / 1K tokens.\footnote{See \url{https://openai.com/pricing} for more details.} We generated around 1,278 GPT-4 summaries for all evaluations in Section~\ref{Sec:eval}. The total cost of API requests was about \$88. 

\section{Quantitative evaluation of hallucination in summaries}
\label{sec: quant-hallucination}
Apart from the qualitative analysis on entity hallucination (Figures \ref{fig:exfic},\ref{fig:exconvall},\ref{fig:exTx}), we conducted a quantitative evaluation on the \corpname{} test set to evaluate whether generated summaries are factually consistent with the source article. We used the $SummaC_{Conv}$  factuality metric (model\_name = `vitc', granularity=sentence-level) from the SummaC model \cite{laban2022summac}, which is an NLI (Natural Language Inference, or more specifically textual entailment) model that is used to measure hallucination based on the assumption that a faithful summary will be entailed by the gold source document. 
Table~\ref{tb: quant-hallucination} shows the $SummaC_{Conv}$ scores ranging from 0 to 1, with 0 indicating the generated summary logically follows from the source document (entailment) and 1 representing that the generated summary contradicts the information in the source document (contradiction).\par
Overall, we can see that adding predicted or gold salient entities to the model significantly improves the factuality of generated summaries and reduces hallucinations (lower scores are better), compare e.g.~the total scores of the CTRL0 (0.737), CTRLPred (0.658) and CTRLGold models (0.537). Among all models, GPTPred produces summaries with the best entailment score (0.222). This demonstrates that adding predicted entities to GPT-4 contributes the most to improving faithfulness. \par
For genre comparison, we found that pre-trained SOTA abstractive summarization models (BRIO and CTRLSum) have higher factuality scores in written genres (e.g.~\textit{fiction}, \textit{textbook}) but not in spoken genres (e.g.~\textit{conversation}). This is unsurprising, as most of the summarization models were trained on written data, whereas spoken data like \textit{conversation} are considered out-of-domain for these models. As such, it is easier for the models to generate summaries that are more factually consistent with the source document (or contain fewer hallucinations) in these ``familiar'' genres. However, similar to our findings in Section~\ref{sec:4.1}, this genre effect does not seem to appear in GPT methods, where spoken genres like \textit{speech} surprisingly outperform written genres like \textit{fiction} and \textit{academic}. As we have indicated in Section~\ref{sec:4.1}, this might be because GPT-4 was trained on a wide variety of data sources, which include political speeches, etc. \par
Interestingly, we found that $SummaC_{Conv}$ scores in \textit{textbook, voyage, and interview} improve the most after adding predicted or gold salient entities to the model (see the scores with \cross[.4pt] in Table~\ref{tb: quant-hallucination}). This indicates that the addition of salient entities is most effective in enhancing faithfulness in these `unusual' genres.

\begin{table*}[h!]
\centering
\resizebox{0.9\textwidth}{!}{%
\begin{tabular}{@{}rrrrrrrr@{}}
\toprule
\textbf{} & \textbf{BRIO} & \textbf{CTRL0} & \textbf{CTRLPred} & \textbf{CTRLGold} & \textbf{GPT0} & \textbf{GPTPred} & \textbf{GPTGold} \\ \midrule
\textbf{ac} & 0.228 & 0.845 & 0.805 & 0.832 & 0.230 & 0.235 & \textcolor{red}{0.349} \\
\textbf{bi} & 0.495 & 0.708 & 0.689 & 0.537 & 0.245 & 0.234 & 0.212 \\
\textbf{cn} & 0.223 & \textcolor{red} {0.885} & \textcolor{red} {0.890} & 0.439\textsuperscript{\cross[.4pt]} & 0.273 & 0.241 & 0.237 \\
\textbf{fc} & 0.233 & \textcolor{blue} {0.300} & \textcolor{blue} {0.355} & 0.367 & 0.228 & \textcolor{red} {0.251} & 0.217 \\
\textbf{it} & 0.543 & 0.862 & 0.714\textsuperscript{\cross[.4pt]} & 0.625 & 0.430 & 0.215\textsuperscript{\cross[.4pt]} & 0.231\textsuperscript{\cross[.4pt]} \\
\textbf{nw} & 0.425 & 0.879 & 0.830 & \textcolor{red} {0.850} & 0.289 & 0.203 & 0.205 \\
\textbf{rd} & 0.255 & 0.840 & 0.851 & 0.739 & 0.268 & 0.242 & 0.217 \\
\textbf{sp} & \textcolor{blue} {0.203} & 0.735 & 0.764 & 0.488 & \textcolor{blue} {0.218} & \textcolor{blue} {0.200} & \textcolor{blue} {0.203} \\
\textbf{tx} & 0.235 & 0.820 & 0.372\textsuperscript{\cross[.4pt]} & \textcolor{blue} {0.337}\textsuperscript{\cross[.4pt]} & 0.338 & 0.208\textsuperscript{\cross[.4pt]} & 0.206\textsuperscript{\cross[.4pt]} \\
\textbf{vl} & 0.248 & 0.439 & 0.484 & 0.412 & 0.245 & 0.232 & 0.231 \\
\textbf{vy} & 0.371 & 0.820 & 0.709 & 0.412\textsuperscript{\cross[.4pt]} & \textcolor{red} {0.545} & 0.201\textsuperscript{\cross[.4pt]} & \textcolor{blue} {0.203}\textsuperscript{\cross[.4pt]} \\
\textbf{wh} & \textcolor{red} {0.553} & 0.719 & 0.435\textsuperscript{\cross[.4pt]} & 0.412 & 0.241 & 0.207 & 0.207 \\\midrule
\textbf{Total} & 0.334 & 0.737 & 0.658 & 0.537 & 0.296 & \textbf{0.222} & 0.226 \\ \bottomrule
\end{tabular}%
}
\caption{$SummaC_{Conv}$ scores on the \corpname{} test set for all systems. The \textcolor{blue}{blue} text shows the highest entailment score across 12 genres and \textcolor{red}{red} text is the highest contradiction score across 12 genres. The best entailment score across all models is \textbf{bolded}. Scores with \cross[.4pt] are the ones that have top 3 $\Delta$ scores in each model compared to the corresponding \textsc{zero} setting. See Table~\ref{tb:1} for genre codes.}
\label{tb: quant-hallucination}
\end{table*}

\section{Detailed summary examples} \label{sec:appendix-qualitative}

\begin{figure*}[h!]
\centering
\includegraphics[width=\textwidth,height=70mm ]{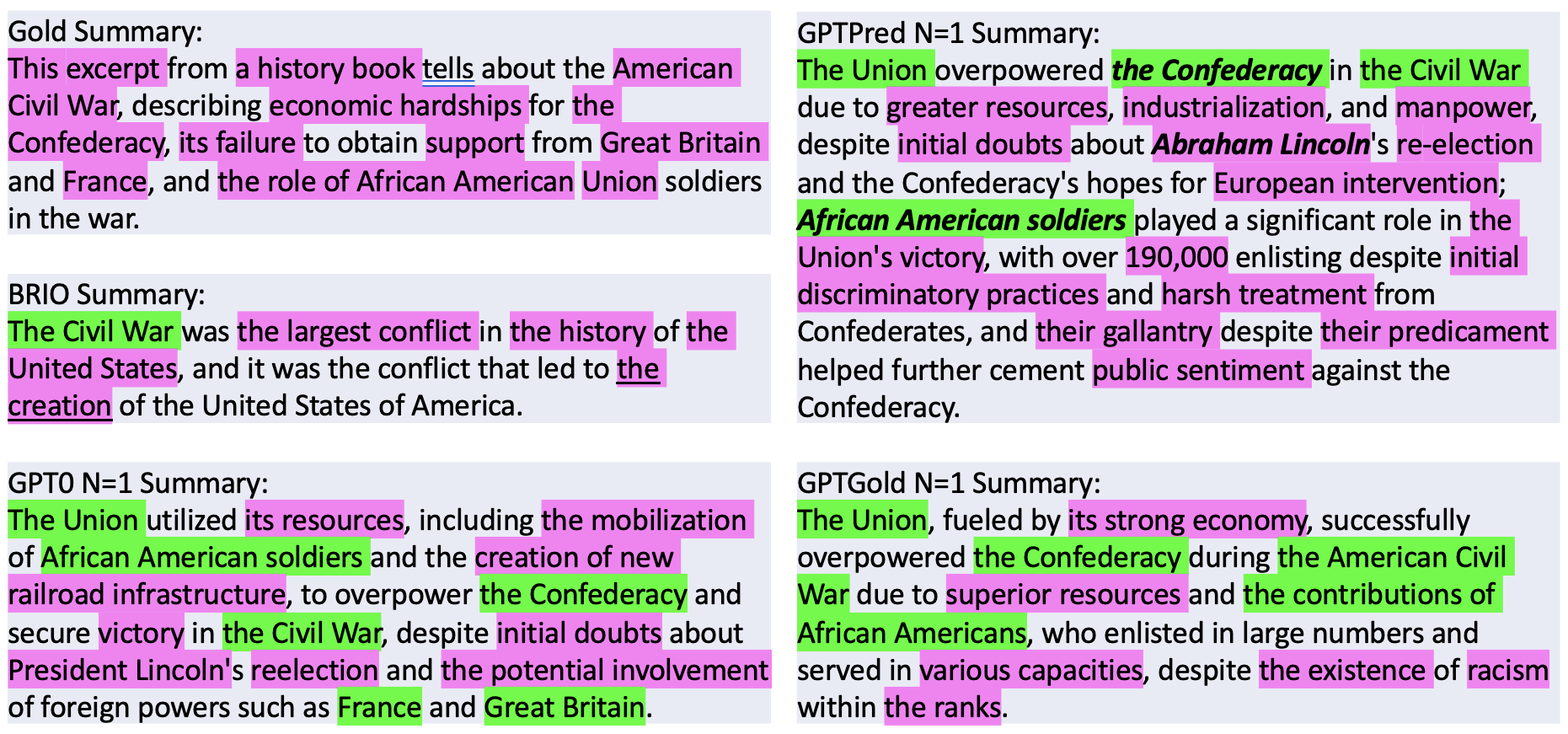}
\caption{An example of generated summaries with BRIO and GPT-4 from a \textit{textbook} document in \corpname{}. The first mention of the entity has been highlighted in pink. Those that have a match in the ground truth summary are highlighted in green. The hallucinated entities are underlined and predicted entities by GPT-4 are in italics.} \label{fig:exTx}
\end{figure*}

Although most of the models perform generally well in written genres, we found that BRIO and GPT models perform unsatisfactorily in the \textit{textbook} genre in summary level evaluation. The possible reasons for this include: (1) The BRIO summaries for \textit{textbook} are too short to include important details of the article and they often contain hallucinated entities. (2) GPTPred summaries contain incorrect salient entities predicted by GPT-4, leading to low ROUGE scores. We found that the headings of textbooks are sometimes misleading. GPT-4 tends to select entities based on their position in the textbook i.e.~entities that are in the beginning or the headings of the textbook article are more likely to be selected as ``salient'', which is not always correct (e.g.~\textit{Abraham Lincoln} is mentioned but never discussed in the underlying document, and the human summary omits his name). See Figure~\ref{fig:exTx} for an example of this. 
\end{document}